\pdfoutput=1

\documentclass[11pt]{article}

\usepackage{acl}
\usepackage{CJKutf8}
\usepackage{booktabs}
\usepackage{times}
\usepackage{soul}
\usepackage{latexsym}
\usepackage{graphicx}
\usepackage{enumitem}
\usepackage{caption}
\usepackage{multirow}
\usepackage[T1]{fontenc}
\usepackage[utf8]{inputenc}
\usepackage{microtype}
\usepackage{inconsolata}
\usepackage{booktabs}

\title{Extract and Attend: Improving Entity Translation in \\ Neural Machine Translation}

\author{
\textbf{Zixin Zeng}$^{1}$\thanks{This work was completed at Microsoft. Corresponding authors: Rui Wang and Xu Tan (\{ruiwa,xuta\}@microsoft.com).}, 
\textbf{Rui Wang}$^{2}$, 
\textbf{
Yichong Leng}$^{3}$, \textbf{Junliang Guo}$^{2}$,\textbf{Xu Tan}$^{2}$,\textbf{Tao Qin}$^{2}$,\textbf{Tie-yan Liu}$^{2}$\\
$^{2}$ Peking University,
$^{2}$ Microsoft Research Asia\\
$^{3}$ University of Science and Technology of China\\
$^{1}$\texttt{1800016623@pku.edu.cn}\\
$^{2}$ \texttt{\{ruiwa,junliangguo,xuta,taoqin,tyliu\}@microsoft.com} \\
$^{3}$\texttt{lyc123go@mail.ustc.edu.cn} 
}

\begin{document}
\maketitle
\begin{abstract}

While Neural Machine Translation~(NMT) has achieved great progress in recent years, it still suffers from inaccurate translation of entities (e.g., person/organization name, location), due to the lack of entity training instances. When we humans encounter an unknown entity during translation, we usually first look up in a dictionary and then organize the entity translation together with the translations of other parts to form a smooth target sentence. Inspired by this translation process, we propose an Extract-and-Attend approach to enhance entity translation in NMT, where the translation candidates of source entities are first extracted from a dictionary and then attended to by the NMT model to generate the target sentence. Specifically, the translation candidates are extracted by first detecting the entities in a source sentence and then translating the entities through looking up in a dictionary. Then, the extracted candidates are added as a prefix of the decoder input to be attended to by the decoder when generating the target sentence through self-attention. Experiments conducted on En-Zh and En-Ru demonstrate that the proposed method is effective on improving both the translation accuracy of entities and the overall translation quality, with up to $35\%$ reduction on entity error rate and $0.85$ gain on BLEU and $13.8$ gain on COMET.

\end{abstract}

\section{Introduction}

%Why to do entity translation
Neural machine translation (NMT) automatically translates sentences between different languages, which has achieved great success~\cite{Bahdanau2015NeuralMT,Sutskever2014SequenceTS, he2016dual,song2019mass,wang2021survey}. Most current works consider to improve the overall translation quality. However, the words in a sentence are not equally important, and the translation accuracy of named entities (e.g., person, organization, location) largely affects user experience, an illustration of which is shown in Table~\ref{tab:importance}. Unfortunately, the translation accuracy of named entities in a sentence is not quite good with current NMT systems~\cite{hassan2018achieving,Lubli2020ASO} due to the lack of training instances, and accordingly more effort is needed.

\begin{table*}[]
    \centering
    \begin{tabular}{c|c}
         \hline
         Source & \begin{CJK*}{UTF8}{gbsn}北岛的绘画展在巴黎地平线画廊开幕。 \end{CJK*}\\
         \hline
         Reference & Bei Dao's painting exhibition opens at Horizon Gallery in Paris. \\
         Output 1 & \textcolor{red}{North Island}'s painting exhibition opens at Horizon Gallery in Paris. \\
         Output 2 & Bei Dao's \textcolor{red}{picture} exhibition opens \textcolor{red}{on} Horizon Gallery in Paris. \\
         \hline
    \end{tabular}
    \caption{Illustration of entity translation in a sentence, where ``\begin{CJK*}{UTF8}{gbsn}北岛\end{CJK*}'' in Chinese can be either a person name or an island. Both outputs 1 and 2 have two different words with red color compared with the reference sentence, while output 2 with correct translation on the entity ``\begin{CJK*}{UTF8}{gbsn}北岛\end{CJK*}'' is much better.}
    \label{tab:importance}
\end{table*}

Recalling the process of human translation, 
when encountering an unknown entity in a sen
tence, humans look up the translation of 
the entity in mental or external dictionaries, and 
organize the translation of the entity together with the translations of other parts to form a smooth target sentence based on grammar and language sense \cite{Gerver1975APA,cortese_1999}. As the original intention of neural networks is to mimic the human brain, the human translation process is also an important reference when dealing with entities in NMT. However, none of the previous works on improving the entity translation in NMT consider both steps in human translation: 1) some works annotate the types and positions of the entities without using the dictionary~\cite{li2018named,modrzejewski2020incorporating}; 2) some works first extract the entity translations from a dictionary~\cite{wang2017sogou} or an entity translation model~\cite{li2018neural,yan2019impact,li2019neural}, and then directly use them to replace the corresponding entities in the translated sentence via post-processing, which only takes the first step of human translation and may affect the fluency of the target sentence; 3) a couple of works use data augmentation or multi-task training to handle the entities in NMT~\cite{zhao2020knowledge,hu2022deep}, which do not explicitly obtain the translation for each entity as the first step in human translation. 

%How we do to solve the problem
Inspired by the human translation process, we propose an Extract-and-Attend approach to improve the translation accuracy of named entities in NMT. Specifically, in the ``Extract'' step, translation candidates of named entities are extracted by first detecting each named entity in the source sentence and then translating to target language based on the dictionary. Considering that some types of entities (e.g. person names) have relatively high diversity and low coverage in dictionaries, we also develop a transliteration\footnote{Transliteration is to convert between languages while keeping the same pronunciation~\cite{karimi2011machine}.} pipeline to handle the entities uncovered by the dictionary. In the ``Attend'' step, the extracted candidates are added to the beginning of the decoder input as a prefix to be attended to by the decoder via self-attention. The Extract-and-Attend approach enjoys the following advantages: 1) the translation candidates of the named entities are explicitly extracted and incorporated during translation, which provides specific 
references for the decoder to generate the target sentence; 2) the extracted candidates are incorporated via self-attention instead of hard replacement, which considers the context of the whole sentence and leads to smooth outputs. The main contributions of this paper are summarized as follows:
\begin{itemize}[leftmargin=*]
    \item We propose to mimic the human translation process when dealing with entities in NMT, including extracting the translations of entities based on dictionary and organizing the entity translations together with the translations of other parts to form a smooth translation.
    \item Accordingly, we propose an Extract-and-Attend approach to improve the quality of entity translation in NMT, which effectively improves the translation quality of the named entities.
    \item Experiments conducted on En-Zh and En-Ru demonstrate that the proposed Extract-and-Attend approach significantly reduces the error rate on entity translation. Specifically, it reduces the entity error rate by up to $35\%$ while also improving BLEU by up to $0.85$ points and COMET up to $13.8$ points.
\end{itemize}

\section{Related Work}

\begin{figure*}[!t]
\centering
\includegraphics[width=\textwidth]{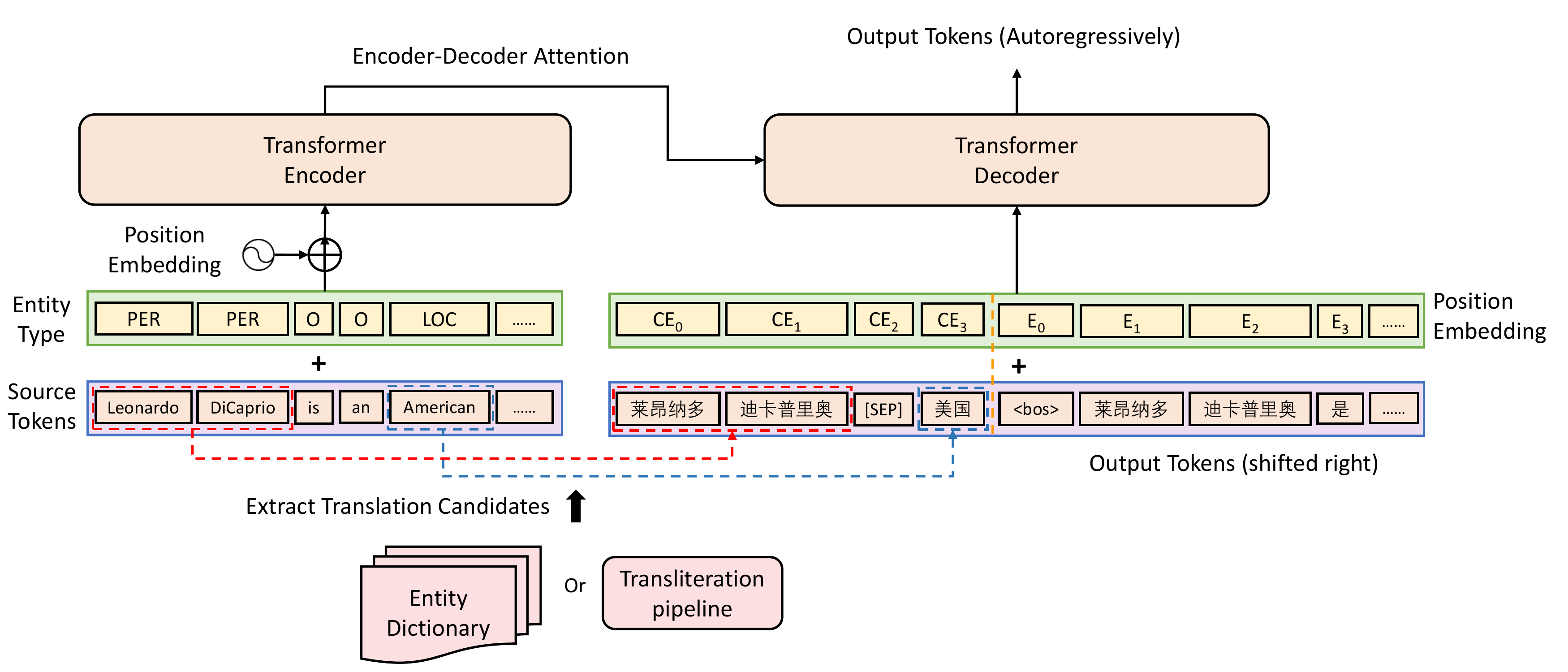}
\caption{Extract-and-Attend approach, where the translation candidates are extracted and added as a prefix of the decoder input. Entity type embeddings are added to the source input (e.g., `PER' for person names, `LOC' for locations and `O' for other tokens other than entities). Independent position embeddings are used for the translation candidates and the shifted output tokens (i.e., `CE' for translation candidates and `E' for output tokens).}
\label{methodpic}
\end{figure*}

To improve the entity translation in NMT, some works focus on annotating named entities to provide type and position information. For example, the inline annotation method~\cite{li2018named} inserts special tokens before and after the entities in the source sentence. The source factor method~\cite{ugawa2018neural,modrzejewski2020incorporating} adds entity type embeddings to the tokens of the entities in the encoder. \citet{xie2022endtoend} attach entity classifiers to the encoder and decoder. One main challenge when dealing with entities is that the entities are quite diverse while the corresponding data is limited compared to the large number of entities. Dictionaries are important supplements to the limited data on entities, which are not utilized in these works. 

With the help of bilingual dictionaries, one common approach to improve the entity translation in NMT is to first extract the translation of source entities based on a dictionary~\cite{wang2017sogou} or an entity translation model~\cite{li2018neural,yan2019impact,li2019neural}, and then locate and replace the corresponding tokens in the target sentence via post-processing. However, such approach only takes the first step of human translation (i.e., extracting the entity translations), since the entity translations are inserted to the target sentence by hard replacement, which affects the fluency of the target sentence. Moreover, this approach is sensitive to the inaccurate predictions made by NER~\cite{modrzejewski2020incorporating}.
%when locating the entities in the target sentence, Named Entity Recognition~(NER)~\cite{li2020survey} is commonly used, which may make inaccurate predictions on the types and positions of the entities~\cite{modrzejewski2020incorporating}. %\hl{Unfortunately, the approach directly replacing the entities in the target sentence is very sensitive to the errors made by NER}. 

Recently, some works take advantage of additional resources (e.g., dictionary) via data augmentation or multi-task training to improve the translation quality on entities. \citet{zhao2020knowledge1} augment the parallel corpus based on paired entities extracted from multilingual knowledge graphs, while DEEP~\cite{hu2022deep} augments monolingual data with paired entities for a denoising pre-training task. The entity translation can also be enhanced by multi-task training with knowledge reasoning~\cite{zhao2020knowledge} and integrating lexical constraints~\cite{wang-2022-integrating}. These methods don't look up translation candidates in bilingual dictionares during inference. Considering that entities are quite diverse, providing specific translation candidates from dictionary may further improve the quality of entity translation.
%hard replace
%knowledge graph & DEEP 

Bilingual dictionaries are also utilized for improving translation quality on rare words or domain-specific terminology. One common approach is to augment training data with pseudo parallel sentences generated based on the dictionary~\cite{zhang-2016-bridging,nag-2020-incorporating,zhao2020knowledge1,peng2020dictionary}. Some works adjust the output probabilities over the vocabulary in the decoder according to the dictionary~\cite{arthur2016incorporating,zhao2018phrase,zhang2021point}. \citet{zhong2020look} attach the definitions of the rare words in the dictionary to enhance the rare word translation. Similarly, \citet{dinu-etal-2019-training} and \citet{exel-etal-2020-terminology} proposed to inject terminology by replacing or inserting translations inline in the source sentence. Though the human translation process when encountering an unknown rare word/terminology or entity is the same, we argue that the two-step human translation process is more suitable for entities. This is because rare words can be polysemous and require context-based disambiguation; on the other hand, each entity is usually linked with a single sense after controlling for entity type. Accordingly, retrieved translations of entities are less ambiguous than other words. On the contrary, domain-specific terminology always has a single sense which has little relevant to context, and thus it is usually with much higher accuracy to identify the terminologies in the domain-specific sentences than entities. Another uniqueness of entities is that some entities are translated by the same rule, which makes it possible to generalize to unseen entities. For example, when translating the names of Chinese people from Chinese to English, Pinyin\footnote{https://en.wikipedia.org/wiki/Pinyin} is commonly used.

\begin{figure*}[!t]
\centering
\includegraphics[width=0.795\textwidth]{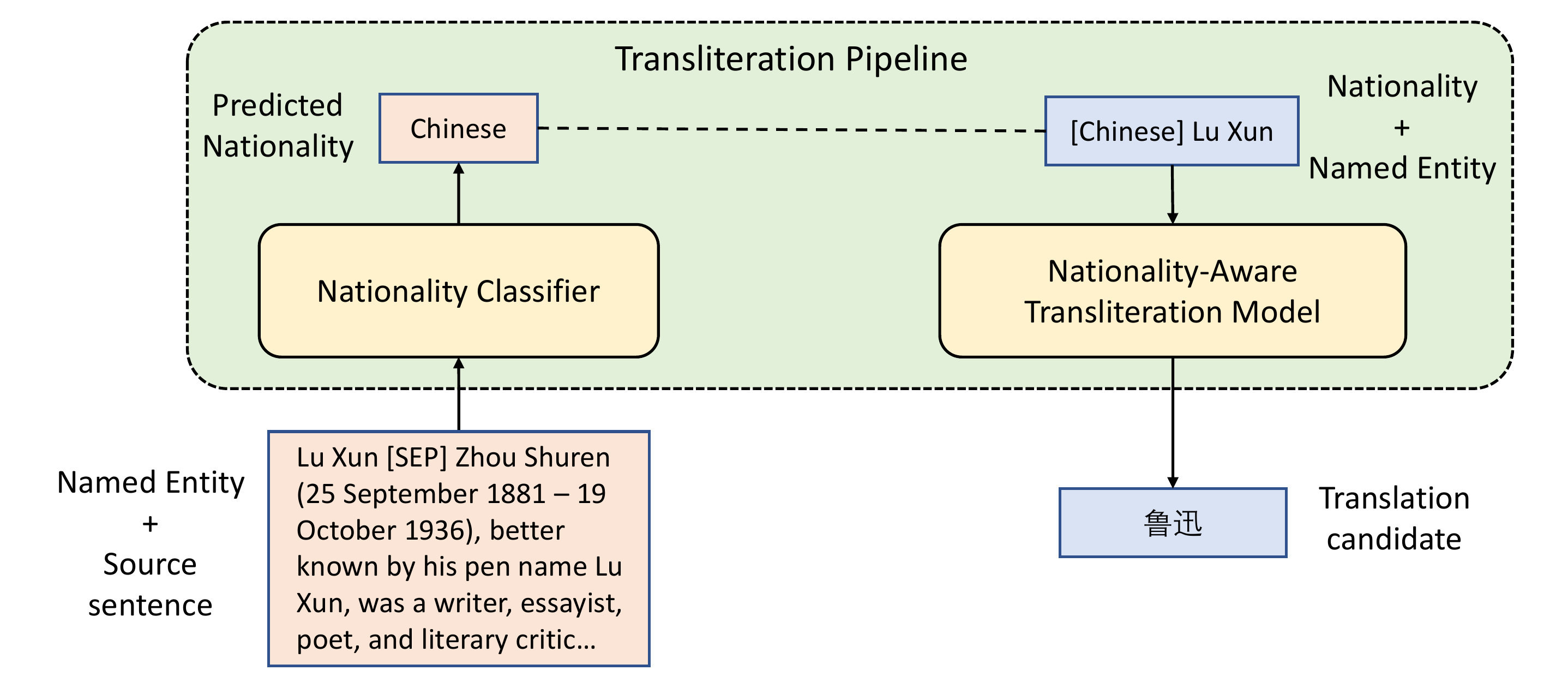}
\caption{Transliteration pipeline.}
\label{transliteration}
\end{figure*}

\section{Improving Entity Translation in NMT}

Inspired by the translation process of humans when encountering an unknown entity, where the translation of the entity is extracted from a dictionary and then organized with the translations of other parts to form a fluent target sentence, we propose an Extract-and-Attend approach. Specifically, we first extract the translation candidates of the entities in the source sentence, and then attend the translation candidates into the decoding process via self-attention, which helps the decoder to generate a smooth target sentence based on the specific entity translations. An overview of the proposed Extract-and-Attend approach is shown in Fig.~\ref{methodpic}, where a Transformer-based~\cite{vaswani2017attention} encoder-decoder structure is adopted. Specifically, to extract the translation candidates, entities in the source sentence are first detected based on NER ~\cite{li2020survey}, then the translation candidates are obtained from a bilingual dictionary. Considering that some types of named entities (e.g., person names) are quite diverse and the coverage in the dictionary of such entities is limited, we also develop a transliteration pipeline to handle entities uncovered by the dictionary. To make the decoder attend to the translation candidates, we add the translation candidates in order as a prefix of the decoder input. In the following sections, we will provide the details of ``Extract'' and ``Attend''.

\subsection{Extracting Translation Candidates}\label{Extract}
Extracting the translation candidates for entities in the source sentence provides explicit references when generating the target sentence in NMT. There are two steps when extracting the entity translation candidates, where the entities in the source sentences are first detected by NER and then translated to the target language. If the entity is found in the bilingual dictionary, we retrieve its translation(s). Although there may be multiple translation candidates for one entity, the entity usually links to a single sense after disambiguating by entity type, and the multiple candidates in the dictionary for one named entity are commonly all correct. For example, ``John Wilson'' can be translated to 
``\begin{CJK*}{UTF8}{gbsn}约翰·维尔逊\end{CJK*}'' or ``\begin{CJK*}{UTF8}{gbsn}约翰·威尔森\end{CJK*}''. During training, we consider the one with shortest Levenshtein distance\footnote{https://en.wikipedia.org/wiki/Levenshtein\_distance} compared to the ground truth translation to encourage the decoder to copy the given candidate. During inference, considering that only the source sentence is available, we select the one with highest frequency in the training set.

The coverage in the dictionary is limited for some types of entities (e.g, person names). Meanwhile, a large number of named entities (e.g., person names and some of locations) are translated by transliteration (i.e., translated according to the pronunciations). Accordingly, we consider to use transliteration to handle such entities if they are uncovered by the dictionary. Transliteration in different countries often follow different rules. For example, names of Chinese persons are transliterated into English via Pinyin, while names of Korean persons are often transliterated via McCune-Reischauer\footnote{https://en.wikipedia.org/wiki/McCune-Reischauer}. Current transliteration models~\cite{kundu2018deep,karimi2011machine,le2019low} do not consider different nationalities for a single language pair, which is an important cause for transliteration errors. Considering this, we develop a nationality-aware transliteration pipeline, which consists of a nationality classifier and a nationality-aware transliteration model. As shown in Fig.~\ref{transliteration}, the nationality classifier takes the source entity and source sentence as input, and predicts the nationality of the entity. Then, the nationality tag is concatenated with the entity and translated by the word-level transliteration model.

\subsection{Attending to Translation Candidates}
We consider to let the decoder attend to the extracted translation candidates via self-attention, which has shown to be more effective in improving entity translation compared to alternative designs (see Section \ref{analysis-attend}). Accordingly, we concatenate extracted candidate translations with ``[SEP]'' and place it before the ``<bos>'' token of the decoder input. In order to identify the alignments between the translation candidates and the corresponding entities in the source sentence, we add entity type embeddings to word embeddings of the entities in the source sentence as \cite{modrzejewski2020incorporating}, and concatenate the corresponding translation candidates in the same order as they are in the source sentence. We demonstrate that our model can correctly align the entities and the corresponding translation candidates in Appendix~\ref{casestudy} via case study. We use independent position embeddings for the translation candidates and the target sentence as shown in Fig.~\ref{methodpic}. The loss on the tokens of translation candidates is ignored. In this way, the decoder can attend to the translation candidates through the attention mechanism in the decoder, which helps improve the performance of the model on translating entities.

\section{Experimental Settings}
In this section, we describe experimental settings including datasets, model configurations, the evaluation criterion and baselines.

\subsection{Datasets}
\label{Dataset}

We conduct experiments on English-Chinese (En-Zh) and English-Russian (En-Ru) translation. We chose language pairs so that the source and target languages come from different scripts\footnote{English uses Latin script, Chinese uses Logographic script, and Russian uses Cyrillic script.}, because cross-script entity translation is more challenging.
Following~\citet{modrzejewski2020incorporating}, three types of named entities are considered, i.e., person name, organization and location. Note that the proposed framework is not limited to the three types and can be applied to other entities (e.g., domain entities).
%by collecting corresponding dictionary or developing entity translation model
\paragraph{Entity dictionary.} Entity pairs and corresponding nationality information are obtained from two multilingual knowledge graphs (i.e., DBPedia and Wikidata). For En-Ru, we extract 401K, 175K and 50K pairs of PER, LOC and ORG entities respectively. For En-Zh, we extract 338K, 200K, 38K pairs of PER, LOC and ORG entities respectively.  Besides, we increase the coverage of the entity dictionary by mining entity pairs from parallel data. First, we use spaCy NER models\footnote{https://pypi.org/project/spacy/} to recognize entities from parallel sentences, then use awesome-align~\cite{dou2021word} to align the source and target tokens and extract the corresponding translations. Infrequent entity pairs or empty alignment results are filtered out. Specifically, we obtain 179K person names, 51K locations, and 63K organizations for En-Ru, and 152K person names, 32K locations, and 39K organizations for En-Zh. 

\paragraph{Dataset for transliteration pipeline.} 
Most person names and part of locations can be translated by transliteration. 
Because the dictionary has relatively high coverage for location entities, we
train the transliteration pipeline based on parallel person names, and use it for both person names and unseen locations. To train the nationality classifier,  we extract English biographies from DBPedia and link them to the entity dictionary, which are translated into Chinese and Russian with custom NMT models. In total, we collect $54$K sentences with person names and nationalities, where $48.2$K, $1.5$K and $3.9$K of them are used as training set, validation set and test set, respectively.  We also merge countries that share the same official language (e.g. USA and UK), and regard the nationalities with fewer than $1000$ examples as ``Other''. For the nationality-aware transliteration model, the paired person names with nationality information from the collected entity dictionary are used. For En-Zh, $316$K, $5$K, and $17$K are used as training set, validation set and test set respectively, and for En-Ru, $362$K, $13$K, $26$K are used as training set, validation set and test set respectively. Besides, we also collect common monolingual person names from various databases\footnote{https://namecensus.com/ \\ http://www.openkg.cn/dataset/cndbpedia \\ https://github.com/wainshine/Chinese-Names-Corpus 
 \\ https://github.com/datacoon/russiannames}, and create pseudo entity pairs via back translation~\cite{BT}. In total, 10K, 1.6M and 560K entities are collected for English, Chinese and Russian respectively.

\paragraph{Dataset for NMT model.} The training data is obtained from UN Parallel Corpus v1.0 and News Commentary Corpus v15\footnote{Available at https://www.statmt.org/wmt20/translation-task.html}. The test data is constructed by concatenating test sets of the WMT News Translation Task (2015-2021) and deduplicating samples. Dataset statistics are shown in Table \ref{wmtdata}. For En-Zh, there are 6.6K PER entities, 4.4K ORG entities and 1.9K LOC entities. For En-Ru, there are 4.9K PER entities, 2.5K ORG entities and 1.2K LOC entities. We use Moses\footnote{https://github.com/moses-smt/mosesdecoder} to tokenize English and Russian corpus, and perform word segmentation on Chinese corpus with jieba\footnote{https://pypi.org/project/jieba/}. We perform joint byte-pair encoding (BPE) by subwordnmt\footnote{https://github.com/rsennrich/subword-nmt} with a maximum of 20K BPE tokens.
\begin{table}[!htbp]\small
\centering
\begin{tabular}{c|c|c}
\hline
Languages & \#Sentences & \#Tokens\\
\hline
En-Ru & 23.5M/25K/13K & 726M/456K/227K \\
En-Zh & 16.2M/20k/19k & 487M/512K/503K \\
\hline
\end{tabular}
\caption{Statistics of NMT datasets (Train/Val/Test).}
\label{wmtdata}
\end{table}

\subsection{Model Configurations and Training Pipeline}
The nationality classifier is fine-tuned from pre-trained BERT checkpoint (base, cased) available on HuggingFace\footnote{https://huggingface.co/bert-base-uncased}. Both the NMT model and the nationality-aware transliteration model use Transformer base architecture~\cite{vaswani2017attention} with $6$-layer encoder and decoder, hidden size as $512$ and $8$ attention heads.

\subsection{Evaluation Criterion and Baselines}
\label{errorrate}
To evaluate the overall translation quality, we compute BLEU and COMET~\cite{rei-etal-2020-comet} scores\footnote{The wmt22-comet-da model is used to calculate COMET scores}.
To evaluate the translation quality on entities, we consider using error rate of entity translation as the evaluation criterion. Following~\citet{modrzejewski2020incorporating}, we evaluate entity error rate by recognizing named entities from the reference sentence, and then checking occurrence in the output sentence, where it is regarded as error if it does not occur. 

We compare our Extract-and-Attend approach with the following  baselines\footnote{The entity resources used in Transformer with Dictionary, Replacement and Placeholder are obtained as is Section \ref{Extract}}:
\begin{itemize}[leftmargin=*]
    \item \emph{Transformer.} The Transformer model is directly trained on parallel corpus.
    \item \emph{Transformer with Dictionary.} 
    The entity dictionary is directly added to the parallel corpus to train a transformer model.
    \item \emph{Replacement.} After identifying entities in the source sentence with NER and aligning them with target tokens, the corresponding tokens are replaced by translation candidates.
    
    \item \emph{Placeholder~\cite{yan2019impact, li2019neural}.} It first replaces the entities in the source sentence with placeholders based on NER and then restores the placeholders in the output sentence with the extracted translation candidates.
    \item \emph{Annotation~\cite{modrzejewski2020incorporating}.} Entity type embeddings are added to the original word embeddings for the tokens of entities in the source sentence.
    \item \emph{Multi-task~\cite{zhao2020knowledge}} It improves the entity translation in NMT by multi-task learning on machine translation and knowledge reasoning.
\end{itemize}

\section{Experimental Results}

\begin{table*}[htbp]\small
\centering
\begin{tabular}{ccccccccc}
\toprule
\multirow{2}{*}{Model} & \multicolumn{2}{c}{$En \rightarrow Ru$} & \multicolumn{2}{c}{$Ru \rightarrow En$} &  \multicolumn{2}{c}{$En \rightarrow Zh$} & \multicolumn{2}{c}{$Zh \rightarrow En$}\\
\cmidrule(lr){2-3}\cmidrule(lr){4-5}\cmidrule(lr){6-7}\cmidrule(lr){8-9}
& BLEU & COMET & BLEU & COMET & BLEU & COMET & BLEU & COMET\\
\midrule
Transformer & 31.83 & 52.2 & 34.63  & 54.0 & 26.32 
 & 34.8 & 27.45 & 41.5 \\
Transformer w/ Dictionary & 31.85 & 53.6 & 34.67 & 56.1 & 26.36 & 38.1 & 27.49 & 43.2\\
Replacement  & 30.52 & 55.2 & 32.01 & 56.7  & 25.92 & 41.4 & 27.21 & 45.0 \\
Placeholder & 31.88 & 57.6 & 34.72 & 59.1 & 26.41 & 42.9 & 27.50 & 47.2 \\
Annotation & 31.91 & 59.4 & 34.84 & 60.5 & 26.44 & 45.8 & 27.73 & 48.0 \\
Multi-task & 31.88 & 57.8 & 34.76 & 60.3 & 26.38 & 45.0 & 27.64 & 47.4 \\
\midrule
Extract \& Attend (ours) & \textbf{32.68} & \textbf{62.2} & \textbf{35.41} &\textbf{63.5 } & \textbf{26.79} & \textbf{ 48.6 } & \textbf{27.98} &\textbf{50.1 }\\
\bottomrule
\end{tabular}
\caption{BLEU and COMET scores on {WMT newstest}. BLEU and COMET scores are statistically higher than baselines across all language pairs with 95\% statistical significance~\cite{koehn-2004-statistical}.}
\label{BLEU}
\end{table*}

\begin{table*}[t] \small
\centering
\begin{tabular}{c|c|c|c|c}
\hline
Model & $En \rightarrow Ru$ & $Ru \rightarrow En$ &  $En \rightarrow Zh$ & $Zh \rightarrow En$\\
\toprule
Transformer & 60.0 & 51.3 & 42.7 & 41.0 \\
Transformer w/ Dictionary & 59.2 & 50.4 &  42.1 & 40.6 \\
Replacement  & 49.6 & 49.8 & 29.5 & 28.9 \\
Placeholder  & 49.7 & 49.3 & 28.6 & 27.9\\
Annotation & 43.2 & 44.5 & 37.4 & 30.0\\
Multi-task & 58.9 & 50.0 & 42.4 & 40.4\\
\midrule
Extract \& Attend (ours) & \textbf{42.7} & \textbf{41.6} & \textbf{27.7} & \textbf{27.5}\\
\bottomrule
\end{tabular}
\caption{Error rates (\%) on {WMT newstest}.}
\label{ER}
\end{table*}

In this section, we demonstrate the effectiveness of the proposed Extract-and-Attend approach by comparing it with multiple baselines. We also conduct experiments to verify the design aspects of ``Extract'' and ``Attend''.

\subsection{Main Results}
\label{mainresult}
BLEU, COMET and entity error rates of the Extract-and-Attend approach with the baselines are shown in Table~\ref{BLEU} and Table~\ref{ER}, where the proposed approach consistently performs the best on all the metrics and language pairs. From the results, it can be observed that: 1) The proposed method reduces the error rate by up to $35\%$ and achieves a gain of up to $0.85$ BLEU and $13.8$ COMET compared to the standard Transformer model; 2) Compared with the annotation method~\cite{modrzejewski2020incorporating}, which annotates the entities in the source sentence based on NER without incorporating any additional resources (e.g., dictionary), the proposed Extract-and-Attend approach takes advantage of the entity dictionary and nationality-aware transliteration pipeline, and reduces the entity error rate by up to $26\%$ while achieving up to $0.77$ points gain on BLEU and $3.0$ points on COMET; 3) Compared with the replacement and placeholder~\cite{yan2019impact,li2019neural} methods, the Extract-and-Attend approach is more robust to NER errors (see \ref{sec:Robustness}) than hard replacement and reduces the error rate by up to $16\%$ while gaining up to $2.1$ BLEU and $7.2$ COMET; 4) Compared to the multi-task~\cite{zhao2020knowledge} method, the Extract-and-Attend approach explicitly provides the translation candidates when decoding, which reduces the entity error rate by up to $35\%$ and improves BLEU by up to $0.8$ points and COMET up to $4.4$ points. We also provide the error rates for different entity types in Appendix~\ref{sec:error-rate-type}, and analyze the effect of dictionary coverage in Appendix~\ref{coverage}

Entity error rates calculated according to Section~\ref{errorrate} may incur false negative errors, which has two main causes. First, as noted by~\citet{modrzejewski2020incorporating}, it is common for NER models to make erroneous predictions. Second, there may be multiple correct translations for one entity, but the ones different from that in the reference sentence are regarded as errors. For example, BMA (British Medical Association) can either be copied in the target sentence, or translated into its Chinese form ``\begin{CJK*}{UTF8}{gbsn}英国医学会\end{CJK*}''. Therefore, we also perform human evaluation wmttest150 (see Table~\ref{en-zh-manual}), where 150 sentence pairs with entities are randomly sampled from the $En \rightarrow Zh$ test set. Compared to automatic evaluation results in Table~\ref{ER}, entity error rates based on human evaluation become lower after eliminating the false negatives, while the relative performance of different models remain almost consistent. Therefore, though there are false negatives in the automatic evaluation as in Section~\ref{errorrate}, it is still a valid metric for evaluating entity translation. Moreover, we observe that the Extract-and-Attend approach performs the best on all three entity types and reduces the total error rate by $32\%$.

\begin{table}[!htbp] \small
\centering
\begin{tabular}{c|c|c|c|c}
\toprule
Model & PER & ORG & LOC & Total\\
\midrule
Transformer &  26.4 & 14.3 & 13.4 & 17.9 \\

Transformer w/ Dictionary & 25.8 & 13.6 & 13.4 & 16.7 \\
Replacement  &  19.8 & 12.4 & 12.6 & 14.8 \\

Placeholder & 18.9 & 12.4 & 10.9 & 13.6 \\

Annotation & 17.9 & 11.4 & 10.9 & 13.3 \\

Multi-task &  21.7 & 12.4 & 12.6 & 15.5 \\

\midrule
Extract \& Attend & \textbf{16.0} & \textbf{11.4} & \textbf{9.2} & \textbf{12.1} \\
(ours) & & & & \\
%\textbf{14.7} & \textbf{17.8} & \textbf{11.0} & \textbf{14.0}\\
\bottomrule
\end{tabular}
\caption{Human evaluation of entity error rates (\%) on wmttest150 for $En \rightarrow Zh$.}
\label{en-zh-manual}
\end{table}

\subsection{Analysis on Extracting}
To investigate the effectiveness of our transliteration pipeline, we implement a variant denoted as Extract-and-Attend (w/o Transliteration), in which we only extract translation candidates covered by the dictionary. From Table~\ref{Transliteration}, we can see that
the translation quality of person names is significantly improved, reducing the error rate by $37\%$;
transliteration is also effective for locations, reducing the error rate by $9\%$. 
Overall, the transliteration model improves BLEU by 0.33 and COMET by 4.1.

\begin{table}[!htbp] \small
\centering
\begin{tabular}{c|c|c|c|c}
\toprule
Model & BLEU & COMET & PER &  LOC \\
\midrule
Extract \& Attend & 26.79  & 48.6 & 25.6 & 31.6 \\
(with Transliteration) & & & \\
Extract \& Attend & 26.46 & 46.5 & 40.8 & 34.8 \\
(w/o Transliteration) & & & \\
%Transformer & 26.32 & 45.3 & 35.7 \\
\bottomrule
\end{tabular}
\captionsetup{font={small}}
\caption{BLEU, COMET and error rates (\%) for $En \rightarrow Zh$.}
\label{Transliteration}
\end{table}

Considering that different transliteration rules may be applied for different countries, we propose to incorporate nationality information during transliteration. To evaluate the effectiveness of utilizing the nationality information in the transliteration pipeline, we compare the performance of the proposed nationality-aware transliteration pipeline with the transliteration model trained on paired entities without nationality information. As shown in Table~\ref{Nation}, adding nationality information during transliteration consistently improves transliteration quality across all language pairs, and is most helpful for $Zh \rightarrow En$, where the transliteration accuracy is improved by $9\%$.
\begin{table}[!htbp] \small
\centering
\resizebox{0.5\textwidth}{8mm}{
\begin{tabular}{c|c|c|c|c}
\toprule
Transliteration & $En \rightarrow Ru$ & $Ru \rightarrow En$ &  $En \rightarrow Zh$ & $Zh \rightarrow En$\\
\midrule
Nationality-aware & 79 & 85 & 95 & 97 \\
w/o Nationality & 74 & 82 & 90 & 88 \\
\bottomrule
\end{tabular}}
\caption{Accuracy of transliteration (\%).}
\label{Nation}
\end{table}

\subsection{Analysis on Attending} \label{analysis-attend}
We also conduct experiments to evaluate the effect of attending translation candidates in the encoder compared to the decoder. Similar to \citet{zhong2020look}, we append translation candidates to the source tokens, where the position embeddings of the translation candidates are shared with the first token of the corresponding entities in the source sentence. Relative position embeddings denoting token order within the translation candidate are also added. As shown in Table~\ref{encoder}, adding the translation candidates to the decoder is better than adding to the encoder. Intuitively, attending to translation candidates in the encoder may incur additional burden to the encoder to handle multiple languages.

\begin{table}[htbp] \small
\centering
\begin{tabular}{c|c|c|c}
\toprule
Model & BLEU & COMET & Error rate \\
\midrule
Extract \& Attend & 26.79 & 48.6 &  27.7 \\
(Decoder) & & \\
Extract \& Attend & 26.56 & 46.2 & 29.8 \\
(Encoder) & & \\
%Transformer & 26.32 & 42.7\\
\bottomrule
\end{tabular}
\captionsetup{font={small}}
\caption{BLEU, COMET and error rates (\%) for $En \rightarrow Zh$.}
\label{encoder}
\end{table}

Some entities have multiple translation candidates in the entity dictionary. To study whether to provide multiple candidates for each named entity, we extract up to three candidates from the entity dictionary. To help the  model distinguish different candidates, we use a separator between candidates of the same entity, which is different from the one used to separate the candidates for different entities. Table~\ref{multiple} shows that adding multiple translation candidates slightly reduces the translation quality in terms of BLEU, COMET and entity error rate. Intuitively, all the  retrieved translation candidates for an entity are typically correct, and using one translation candidate for each entity provides sufficient information.
\begin{table}[htbp] \small
\centering
\begin{tabular}{c|c|c|c}
\toprule
Model & BLEU & COMET & Error rate \\
\midrule
Extract \& Attend & 26.79 & 48.6 & 27.7\\
(single candidate) & & \\
Extract \& Attend & 26.75 & 47.9 & 27.8\\
(multiple candidates) & & \\
\bottomrule
\end{tabular}
\captionsetup{font={small}}
\caption{BLEU, COMET and error rates (\%) for $En \rightarrow Zh$.}
\label{multiple}
\end{table}

\subsection{Inference Time}\label{infer-time}
Extracting translation candidates requires additional inference time, including the delays from NER and transliteration pipeline. Specifically, the average inference time for standard Transformer, Replacement, Placeholder, Annotation, Multi-task and our method are 389ms, 552ms, 470ms, 416ms, 395ms, 624ms\footnote{Evaluated on a P40 GPU with batch size  of 1, other experimental settings same as Section 4}.

\section{Conclusion}
In this paper, we propose an Extract-and-Attend approach to improve the translation quality in NMT systems. Specifically, translation candidates for entities in the source sentence are first extracted, and then attended to by the decoder via self-attention. Experimental results demonstrate the effectiveness of the proposed approach and design aspects. Knowledge is an important resource to enhance the entity translation in NMT, while we only take advantage of the paired entities, nationality and biography information. In the future work, it is interesting to investigate how to make better use of the knowledge, which can be obtained from knowledge graphs and large-scale pre-trained models. Besides, the proposed Extract-and-Attend approach also has some limitations. First, our method requires additional entity resources, which may be difficult to obtain for certain language pairs. With the development of multilingual entity datasets like Paranames~\cite{saleva2022paranames}, we are optimistic such resources will be more accessible in the near future. Second, as demonstrated in Section~\ref{infer-time}, extracting translation candidates increases inference time. Due to space limitation, more limitations are discussed in Appendix \ref{append-limit}.

\bibliography{anthology,custom}
\bibliographystyle{acl_natbib}

\begin{table*}[!t]
\centering
\begin{tabular}{cp{12cm}}
\hline
\textbf{Source} & 
\underline{Simone} , \underline{Gabby} and \underline{Laurie} all took the same path as \underline{Aly} and \underline{Madison}  to make the \underline{Olympic} team .\\
\textbf{Reference} & \begin{CJK*}{UTF8}{gbsn}\underline{西蒙}、\underline{加布丽埃勒}和\underline{劳瑞}进入\underline{奥运}代表队的途径跟\underline{阿里}及\underline{麦迪逊}一样。\end{CJK*}\\
\textbf{Baseline}  &
\begin{CJK*}{UTF8}{gbsn}\underline{Simone}、\underline{Gabby}和\underline{劳瑞}进入\underline{奥运}代表队的途径跟\underline{Aly}及\underline{麦迪逊}一样。\end{CJK*}\\
\textbf{Ours} & \begin{CJK*}{UTF8}{gbsn}\underline{西蒙}、\underline{加比}和\underline{劳瑞}进入\underline{奥运}代表队的途径跟\underline{阿里}及\underline{麦迪逊}一样。\end{CJK*}\\
\hline
\textbf{Source} & \underline{Lomachenko} defends his belt against \underline{Miguel Marriaga} on Saturday night at 7 on \underline{ESPN} .\\
\textbf{Reference} & \begin{CJK*}{UTF8}{gbsn}在 周六晚上7点的\underline{ESPN}比赛中,\underline{洛马琴科}战胜了\underline{米格尔·马里亚加}，保全了他的地位。 \end{CJK*}\\
\textbf{Baseline}  &
\begin{CJK*}{UTF8}{gbsn}\underline{Lomachenko} 在周六晚上7点在\underline{ESPN}上为 \underline{Miguel Marriaga}辩护。\end{CJK*}\\
\textbf{Ours} & \begin{CJK*}{UTF8}{gbsn}\underline{洛马琴科}周六晚7点在\underline{ESPN}对阵\underline{米格尔-玛利亚加}的比赛中卫冕他的腰带。\end{CJK*}\\
\hline
\textbf{Source} & \underline{iCloud} ’ s main data center at \underline{Gui-An New Area} will be the first data center \underline{Apple} has set up in \underline{China} . On completion , it will be used to store the data of \underline{Apple} users in \underline{China} .
 \\
\textbf{Reference} & \begin{CJK*}{UTF8}{gbsn} \underline{iCloud贵安新区}主数据中心也将是\underline{苹果公司}在\underline{中国}设立的第一个数据中心 项目，项目落成后，将用于存储\underline{中国苹果} 用户的数据 。\end{CJK*}\\
\textbf{Baseline}  & 
\begin{CJK*}{UTF8}{gbsn}\underline{iCloud}在\underline{桂安新区}的主要数据中心将是\underline{苹果}在\underline{中国}建立的第一个数据中心。完成后，它将用于存储\underline{中国苹果}用户的数据。\end{CJK*}\\
\textbf{Ours} & \begin{CJK*}{UTF8}{gbsn}\underline{iCloud}在\underline{贵安新区}的主要数据中心将是\underline{苹果}在\underline{中国}建立的第一个数据中心。完成后，它将用于存储\underline{中国苹果}用户的数据。\end{CJK*}\\
\hline

\end{tabular}
\caption{\mbox{Examples of $En \rightarrow Zh$ entity translation. Entities are underlined.}}
\label{cases}
\end{table*}

\appendix
\section{Appendix}
\label{sec:appendix}
\subsection{Case Study}
\label{casestudy}
We also conduct a case study on the $En \rightarrow  Zh$ test set to demonstrate the capability of our model when handing multiple entities in a sentence. As shown in Table~\ref{cases}, the outputs of our model normally has correct alignments between the translations and the corresponding entities in the source sentence. Besides, the baseline model has a strong tendency to copy unfamiliar entities in the source sentence, while our model can alleviate this problem and encourage the translation model to incorporate proper transliteration.

\subsection{Error Rates by Entity Type}
\label{sec:error-rate-type}
To alleviate the problem of false errors caused by NER, We aggregate across all language pairs and calculated the average error rate for each type of entity. From Table \ref{error-rate-entity-type}, it is shown that our method outperforms all baselines for PER, ORG and LOC entities. 
\begin{table}[!htbp]
\centering
\begin{tabular}{c|c|c|c}
\toprule
Model & PER & ORG & LOC \\
\midrule
Transformer & 50.4 & 42.4 & 37.5 \\
Transformer w/ Dictionary & 49.8 & 41.7 & 37.2 \\
Replacement & 35.2 & 38.9 & 35.2 \\
Placeholder & 34.5 & 39.0 & 33.9 \\
Annotation  & 35.7 & 40.2 & 34.1 \\
Multi-task  & 49.2 & 41.4 & 37.6 \\
ours        & 29.9 & 38.1 & 33.4 \\
\bottomrule
\end{tabular}
\caption{Error rates (\%) on {WMT newstest} by entity type.}
\label{error-rate-entity-type}
\end{table}

\subsection{Robustness against NER errors}
\label{sec:Robustness}
To test the robustness against NER errors,
% of our method and baselines, 
we filter the samples in which incorrect candidates are collected, which can result from NER errors and transliteration errors. Compared to the Transformer baseline, in 32\% of the cases, the extract and attend method is misguided by the incorrect candidates, while for the replacement and placeholder approaches 100\% of the cases is misguided. Accordingly, our method is arguably more robust against NER errors.

\subsection{Analysis of Dictionary Coverage}\label{coverage}
To analyze the performance of our approach on domains not well covered by the dictionary, we evaluate our approach and baselines on OpenSubtitles dataset~\cite{Lison2016OpenSubtitles2016EL}. Because there is no official test set for this dataset, we randomly sample 10K En-Zh sentence pairs. There are 3.6K PER entities, 1.1K ORG entities and 1.1K LOC entities in this test set. Compared to the dictionary coverage of 32.4\% for WMT newstest, the dictionary coverage is only 15.2\% for the Opensubtitles test set. The overall entity error rates are shown in Table \ref{opensub}. Our results show that even when the coverage of the entity dictionary is relatively low, the proposed Extract-and-Attend framework achieves consistent improvement in entity error rates compared to alternative methods. 

\begin{table}[!htbp]
\centering
\begin{tabular}{c|c}
\toprule
Model & Error Rate(\%) \\
\midrule
Transformer & 29.6 \\
Transformer w/ Dictionary & 29.2\\
Replacement &   26.8 \\
Placeholder &   26.3\\
Annotation  &  27.9 \\
Multi-task  & 28.2 \\
\textbf{ours}        &  \textbf{24.9} \\
\bottomrule
\end{tabular}
\caption{Entity error rates (\%) on OpenSubtitles test set for $En\rightarrow Zh$.}
\label{opensub}
\end{table}

\subsection{Comparison with VecConstNMT}
\label{sec:VecConstNMT}
Some researchers have proposed VecConstNMT to mine and integrate lexical constraints from parallel corpora, which can potentially improve entity translation quality~\cite{wang-2022-integrating}.  We compare our method with VecConstNMT on $En \rightarrow Zh$ and $Zh \rightarrow En$. For $En \rightarrow Zh$, and the results are shown in Table~\ref{Error_VecConstNMT}. Possible reasons that our method outperforms their method include: (1) our method uses additional resources such as dictionaries (2) a relatively small portion of lexical constraints are related to entity translation. 
\begin{table}[!htbp]
\centering
\begin{tabular}{c|c|c}
\hline
Model & $En \rightarrow Zh$ & $Zh \rightarrow En$ \\
\hline
VecConstNMT  & 31.8 & 28.1 \\
ours        & 27.7 & 27.5 \\
\hline
\end{tabular}
\caption{Error rates (\%) on {WMT newstest}.}
\label{Error_VecConstNMT}
\end{table}

\subsection{Extended discussion of limitations}
\label{append-limit}
Though errors caused by NER are alleviated by attending to the translation candidates via self-attention, the quality of the extracted translation candidates is still affected by NER accuracy and dictionary coverage, and higher quality of translation candidates normally leads to better performance. Another issue worth noting is the evaluation criterion for entity translation. As mentioned in Section~\ref{mainresult}, automatically calculating the error rate on entities based on NER and the reference sentence incurs false negative errors, and better criteria to evaluate the translation quality of entities are needed. What's more, in this paper we assume transliteration rules are the same for regions using the same language and assume that nationality is the same as language of origin, which may be inappropriate in some rare cases. Last but not least, considering that languages may have their own uniqueness,  experiments on other language pairs are still needed.

\end{document}